\documentclass[]{eptcs}
\providecommand{\event}{FMAS 2023}
\def\titlerunning{Towards Formal Fault Injection}
\def\authorrunning{A. Farooqui \& B. Sangchoolie}
\usepackage[utf8]{inputenc}
\usepackage{enumitem}
\usepackage{graphicx}      
\usepackage{adjustbox}

\usepackage{iftex}

\ifpdf
  \usepackage{underscore}         
  \usepackage[T1]{fontenc}        
\else
  \usepackage{breakurl}           
\fi
\usepackage{booktabs}

\usepackage{listings}
\usepackage{comment}
\usepackage[T1]{fontenc}
\usepackage{braket}
\usepackage{multirow}
\usepackage{titlesec}
\usepackage{pgf}
\usepackage{tikz}
\usepackage{multirow,bigdelim}
\usetikzlibrary{mindmap,trees}

\usepackage{smartdiagram}

\usepackage{amsmath,amssymb,amsfonts}
\usepackage{graphicx}
\usepackage{textcomp}
\usepackage{xcolor}
\usepackage{multirow} 
\definecolor{Gray}{gray}{0.9}  
\usepackage{caption} 
\usepackage{color, colortbl}  
\usepackage[english]{babel}
\usepackage{amsmath}
\usepackage{algpseudocode}  
\usepackage{multicol} 
\usepackage{dirtytalk}  
\usepackage{soul} 

\title{  
Towards Formal Fault Injection for Safety Assessment of Automated Systems
\thanks{This work was partly supported by the VALU3S project, which has received funding from the ECSEL Joint Undertaking (JU) under grant agreement No 876852. The JU receives support from the European Union’s Horizon 2020 research and innovation programme and Austria, Czech Republic, Germany, Ireland, Italy, Portugal, Spain, Sweden, Turkey. This work has also been partly financed by the CyReV project, which is funded by the VINNOVA FFI program – the Swedish Governmental Agency for Innovation Systems (Diary number: 2019-03071). }
}

\author{Ashfaq Farooqui, Behrooz Sangchoolie
\institute{Dependable Transport Systems, RISE Research Institutes of Sweden, Borås, Sweden}
\email{\{ashfaq.farooqui, behrooz.sangchoolie\}@ri.se }
}

\begin{document}

\maketitle

\begin{abstract}
Reasoning about safety, security, and other dependability attributes of autonomous systems is a challenge that needs to be addressed before the adoption of such systems in day-to-day life. \emph{Formal methods} is a class of methods that mathematically reason about a system's behavior. Thus, a correctness proof is sufficient to conclude the system's dependability. However, these methods are usually applied to abstract models of the system, which might not fully represent the actual system. \emph{Fault injection}, on the other hand, is a testing method to evaluate the dependability of systems. However, the amount of testing required to evaluate the system is rather large and often a problem. 
This vision paper introduces \emph{formal fault injection}, a fusion of these two techniques throughout the development lifecycle to enhance the dependability of autonomous systems. We advocate for a more cohesive approach by identifying five areas of mutual support between formal methods and fault injection. By forging stronger ties between the two fields, we pave the way for developing safe and dependable autonomous systems. This paper delves into the integration's potential and outlines future research avenues, addressing open challenges along the way.

\end{abstract}

\section{Introduction}

Safety- and security-critical systems continue to be integrated into our daily lives. 
Ensuring the safety and security is a multi-disciplinary challenge, where design, development, and evaluation play a crucial role. Thus a strong emphasis on updating current engineering practices to create an end-to-end  verification and validation process that integrates all safety and security concerns into a unified approach is key to adopt these systems~\cite{koopman2017autonomous}. Already, \emph{formal methods} and \emph{fault injection} are used in different parts of the development lifecycle to ensure the system is safe and dependable.

\emph{Formal methods} refers to mathematically rigorous techniques for specifying and verifying software and hardware systems. 
To many researchers, the necessity of formal methods is now a given~\cite{Woodcock_Larsen_Bicarregui_Fitzgerald_2009}. However, this has yet to be the case from an industrial perspective.
Several reasons have been suggested for this situation, including a lack of accessible tools, high costs, incompatibility with existing development techniques, and the fact that these methods require a certain level of mathematical sophistication~\cite{knightChallengesUtilizationFormal1998b,knightWhyAreFormal1997a,c.michaelWhyEngineersShould1997}.

\emph{Fault injection(FI)}, on the other hand, is an established method used for the measurement, test, and assessment of dependable computer systems in extreme stress or faulty conditions. Functional safety standards such as IEC~61508~\cite{iec61508} and ISO~26262~\cite{iso26262} recommend the use of FI to prove that malfunctions in electrical and/or electronic systems will not lead to violations of safety requirements.
In comparison to formal methods, FI could be done after the complete system is built. 
This way, FI could be used to study the impact of a fault in one system and its propagation and impact in the complete end to end system.
FI is also used to evaluate security properties of computer systems by means of \textit{attack injection}. Avizienis et al.~\cite{avizienis2004}, consider an attack to be a special type of fault which is human made, deliberate and malicious, affecting hardware or software from external system boundaries and occurring during the operational phase.

Formal methods for safety analysis and fault injection are complementary techniques, and the choice of approach depends on the specific system being evaluated and the goals of the evaluation. Each of these methods has its benefits and shortcomings.  While formal methods are commonly employed  in the early design phase, fault injection testing is performed towards the later stages of development, where the system or its simulation exists. Unfortunately, the knowledge gained from the formal methods at the design phase is rarely reused in other development lifecycle phases when conducting fault injection experiments. Conversely, feedback from fault injection analysis is rarely used to improve the formal design of the system. The underlying problem is the need for common semantics and knowledge sharing across the different communities.
It is clear that to deal with modern autonomous systems, formal methods and fault injection will have to be integrated in a smart way to be able to specify, verify, and validate systems, and be understandable by people without a background in formal methods. This last point is essential for autonomous systems, where they must be certified before they can be used.

By looking at the current state of applied research within both fields 
this paper introduces \emph{formal fault injection}  to help develop dependable autonomous systems.

\section{Integrating Formal Methods and Fault Injection}
\label{sec:integrationFIFM}

The following section highlights the different research directions leveraging the existing state-of-the-art. Additionally, we highlight the new possibilities that will open up as a consequence of this integration and our research plans.

\begin{enumerate}

\item{\textbf{Design level fault analysis:}}
Some studies focus on analyzing fault impacts during the design phase, using formal methods. Automating Failure Mode and Effects Analysis (FMEA) and Fault Tree Analysis (FTA) for system safety analysis~\cite{Bozzano_Cimatti_Mattarei_Tonetta_2014,Selvaraj_Fei_Fabian_2020,ORTMEIER2007139}, are well-established for within the formal community. Other studies concentrate on developing control strategies to safeguard systems against cyber-security attacks~\cite{Su_2018,Hadjicostis_Lafortune_Lin_Su_2022,Meira-Goes_Kang_Kwong_Lafortune_2020}. These approaches primarily target system design and are implemented during early development phases. Despite the industry standards of FMEA and FTA techniques, integrating  formal methods into the development cycle has encountered limited adoption, mainly due to the scarcity of practical tools. Furthermore, these methods often face computational challenges, restricting their applicability to only a small portion of the system. Consequently, they typically address a fixed set of faults. It is valuable to explore how insights from the fault injection phase could be incorporated at a more generic abstraction level to streamline analysis process early in the design phase.

\item{\textbf{Learning the behavior of the faulty system via model learning/learning-based testing:}}

Model learning~\cite{Farooqui_2021,fengSstar,meinke2013lbtest} seeks to devise techniques for acquiring discrete formal models of systems by observing or interacting with them. These techniques engage in iterative testing of the system, or its simulation, to progressively learn the behavior. Model learning techniques are often applied in combination with other tools and methodologies: model checking~\cite{cassel2016active} for model verification, testing methods~\cite{meinke2013lbtest} to rigorously assess system behavior, and supervisory synthesis~\cite{Farooqui_2021} to derive supervisory controllers for system control.

we envision integrating fault analysis into the model learning phase. This analysis approach offers insights into a system's fault-handling capabilities. The resultant model encompasses both nominal and faulty behaviors, enabling offline safety  analysis. Moreover, these models hold the potential to serve as authoritative proof of system safety, offering a resource for regulatory authorities. 

Existing model learning tools, already applied in select industrial  contexts~\cite{mides,isberner2015open}, provide interfaces external systems. However, an investigation is warranted to devise methods for introducing fault models into these tools and subsequently evaluating their utility. An overarching challenge lies in the scale of the resulting models. Nominal models themselves often attain considerable complexity, leading to challenges associated with state-space explosion. Augmenting these models with fault scenarios is sure to encounter state-space limitations, even for relatively modest systems.

\item{\textbf{Using formal models for reducing the fault space:}}
Executing exhaustive fault injection campaigns 
is not practical. In most cases, such an approach would result in executions that do not contribute significantly to safety analysis. The challenge lies in identifying the optimal set of test instances that effectively analyze a system's dependability. Numerous testing methods have been proposed to address this challenge, including probabilistic approaches~\cite{Jha_Banerjee_Tsai_Hari_Sullivan_Kalbarczyk_Keckler_Iyer_2019}, coverage-based techniques~\cite{Cukier_Powell_Ariat_1999}, and heuristic as well as machine learning-based methods~\cite{Moradi_Oakes_Saraoglu_Morozov_Janschek_Denil_2020, Moradi_Oakes_Denil,Sangchoolie2022,Sedaghatbaf2022}. However, most of these methods rely on some level of prior knowledge about the target system. The availability of such knowledge poses limitations in practical applications~\cite{Moradi_Oakes_Denil, Jha_Banerjee_Tsai_Hari_Sullivan_Kalbarczyk_Keckler_Iyer_2019}.

Yet, when formal specifications exist for a specific system, they inherently contain valuable information that can be leveraged. These specifications offer insights into critical faults and their configurations, which are most likely to lead to system failures. This knowledge about fault configurations proves invaluable when designing fault campaigns. Unfortunately, such utilization is often overlooked.

To address this gap, we propose an investigation into the development of common semantics that can harmonize formal specifications and fault injection techniques. Additionally, we recommend the creation of tools to facilitate the seamless integration of results from both methods. This integration holds the potential to enhance the effectiveness of fault analysis while leveraging the rich information contained within formal specifications.

\item{\textbf{Falsification for fault analysis:}}
Falsification methods typically come into play once the system has been implemented, typically in the later stages of the development lifecycle. Since falsification and fault injection share similar approaches and setups, integrating both these methods represents one of the most straightforward way towards a formal fault injection analysis. Both these methods operate with limited knowledge of the system under test. Falsification primarily focuses on testing the input space, while fault injection adopts a broader perspective by also assuming the presence of fault(s) within the system. 

Several generalized methods and tools have proven effective in the formal community for these purposes.
For instance, tools like Scenic and VerifyAI~\cite{Fremont_Kim_Dreossi_Ghosh_Yue_Sangiovanni-Vincentelli_Seshia_2020,Dreossi_Fremont_Ghosh_Kim_Ravanbakhsh_Vazquez-Chanlatte_Seshia_2019} take a probabilistic approach to generate scenarios intelligently and test the system. Tools like Breach~\cite{Donze_2010} and HyConf~\cite{Aerts_Mousavi_Reniers_2015} interface with MATLAB/Simulink models for falsification. In the realm of fault injection, AV-fuzzer~\cite{Li_Li_Jha_Tsai_Sullivan_Hari_Kalbarczyk_Iyer_2020} aligns closely with falsification approaches.

While most falsification approaches aim to find input values that lead to violations of the system's specifications, they typically do not distinguish between valid and faulty inputs. To assess a system's safety, falsification engines can be employed to identify boundaries within the state-space. Subsequently, fault injection analysis focuses on these boundary values and faulty inputs to analyze the impact of faults. In this context, exploring techniques to augment the falsification engine with both nominal and faulty behavior represents a promising avenue for further investigation.

\item{\textbf{A formal specification language for fault injection:}}
To bridge the envisioned integration of formal methods and fault injection into practical application, a critical missing element is a well-defined~\emph{formal specification language} that can act as an interface between the two domains. Therefore, it becomes imperative to delve into the realm of formal specification languages from both theoretical and practical standpoints. A precedent is set by Bessayah et al.~\cite{Bessayah_Cavalli_Martins_2009}, who successfully demonstrated the use of Hoare Logic~\cite{Hoare_1969} as a specification language for implementing fault injection in communication systems. However, there have been limited efforts to evaluate the suitability of such a language for autonomous systems. Hence, we propose a comprehensive exploration of available formal specification languages to assess their compatibility with fault injection methodologies and to pinpoint areas of research inquiry apart from studies to develop such a language.

\end{enumerate}

\section{Formal Fault Injection}

In the past few years, a paradigm called “shift-to-left” has inspired researchers to go towards simulation-based and model-based verification and validation of automated systems. The rise of simulation-based development is a key reason we believe this to be the right time to start investigating the integration of formal methods and fault injection-based testing. 
Notably, simulation-based methodologies have firmly established themselves in both the formal methods and fault injection communities. This shared foundation provides a common ground for the implementation of the proposed integration methods.

\begin{figure}[h]  \centering
\begin{minipage}[h]{0.4\textwidth} \centering
\smartdiagramset{ distance planet-satellite=2.3cm} 
\adjustbox{scale=0.5}{
\smartdiagram[connected constellation diagram]{Formal Specification\\Language, Design level\\ fault analysis,Continuous\\model learning,Falsification based\\ traditional FI }\
}
    \caption{Conceptual overview of the proposed methodology.}
    \label{fig:ffi}
    \end{minipage}
\begin{minipage}[h]{0.4\textwidth} 
\smartdiagramset{
uniform arrow color=true,
arrow color=gray!50!black,
}
\adjustbox{scale=0.5}{\smartdiagram[flow diagram:horizontal]{Design, Implement,Test}    }
\caption{Generalized overview of the development lifecycle.}
    \label{fig:developmentCycle}
    \end{minipage}
\end{figure}

Figure~\ref{fig:ffi} provides a birds-eye view of the proposed integration. Several methodologies exist in literature and practice that define the development lifecycle, such as  V-method~\cite{Graessler_Hentze_2020}, Waterfall~\cite{Rovce}, and Agile~\cite{Reifer_2002}, to name a few. Most of these methodologies include the three phases: design, implementation, and testing phases, in an iterative manner and are depicted in Figure~\ref{fig:developmentCycle}. This cycle of development is valid at various abstraction levels of the product lifecycle from the initial conceptual design, feature development,  simulation and the final product development.
During each phase, a particular set of methods as discussed in Section~\ref{sec:integrationFIFM} can be mapped to the phases and are depicted using the similarly colored bubbles in Figure~\ref{fig:ffi}.
These different phases are never isolated and are continuously updated with feedback from one another. The formal specification language makes it possible for this feedback to be easily translatable between the different phases. 

It is crucial to recognize that there is no universal solution applicable to all scenarios. Therefore, the proposal does not revolve around creating a singular specification language and its corresponding toolkit. Instead, it presents a high-level methodology for evaluating and constructing dependable systems. It acknowledges that specific system characteristics may demand different formalisms. Hence, the idea is to establish a collection of formal specification languages, each supported by its toolset, to facilitate this methodology. Furthermore, these languages, tools, and application approaches can vary across industries, necessitating a multifaceted strategy to address existing limitations and explore new possibilities. Above all, the aim is to establish a consistent and reproducible framework for assessing system dependability.

The assertion of a system's dependability must always be substantiated by the possibility of reproducing the results of all conducted tests. By formally specifying the entire injection methodology, it becomes possible to perform analysis, and potential replication or extension of the results by interested parties. The overarching vision of this work is to enable a standardized interface for all stakeholders involved in ensuring system safety. For instance, developers and companies can employ formal proofs to demonstrate a product's dependability, governmental certification agencies can utilize available data to enhance certification processes, and third-party auditors can scrutinize systems from security and safety perspectives using existing information. Aligning different phases of the development lifecycle with a common language paves the way for formalizing safety evaluations.

Furthermore, ongoing national and international efforts within the autonomous driving domain aim to define and develop a safety assurance framework~\cite{sunrise-europe.eu,pegasus-EN,sakura} for verification and validation of autonomous systems. These methods aim to develop a database of testing scenarios to validate a  system.   Therefore, in addition to developing tools and techniques, we propose investigating the feasibility of recommending formal fault injection as a best practice through responsible standardization organizations.

\section{Insights from early experiments}

In this section we share our initial experiences and insights from applying formal techniques in simulation based fault injection. Although these experiences are common within the formal community, we believe they offer valuable insights to those interested in bridging the formal and fault injection domains.
\subsection{The case study}
Maleki and Sangchoolie~\cite{Maleki2021} investigated the effects of faults on Advanced Driver Assistance Systems using the Simulation of Urban Mobility (SUMO)~\cite{Alvarez_Lopez_Microscopic_Traffic_Simulation}. We use this work as a basis to study the integration of formal techniques and fault injection.
The scenario used in that work~\cite{Maleki2021} revolves around a three-lane road, spanning 750 meters. Two vehicles, a leader and a follower, navigate this road. The overarching requirements mandate that these vehicles not collide, successfully traverse the road, and maintain a speed not exceeding 36 m/s--the maximum permissible speed. Furthermore, these requirements should endure even when additional vehicles are introduced to the scenario, thus preserving safety and functionality amidst traffic.

To enhance the above with formal methods, we explore the following approaches.
\begin{itemize}
    \item{\emph{Utilizing SAT Techniques for Vehicle Controller Modeling:}} This approach involves modeling the vehicle models from SUMO into a SAT solver. By doing so, we enable the solver to identify potential (faulty) parameters that could lead to the violation of requirements. This approach effectively narrows down the fault space that needs to be tested. Subsequently, these identified inputs can be verified within SUMO to assess their impact on the system's behavior.
    \item{\emph{Applying Model Learning for Faulty Model Generation:}} Here, we connect SUMO to a model learning tool controlled over TCP/IP and allow the system to learn an abstracted model that closely describes the behavior of the simulation.
\end{itemize}

\subsection{Insights} 
Below we provide some of our insights from early experiments on the previously mentioned case study. 
\begin{enumerate}
    \item{\textbf{Finding suitable abstractions:}} One of the most significant takeaways from our study was the challenge of achieving a suitable abstraction level that aligns  with the  use case. As the system was implemented within the SUMO framework, we encountered limitations in deriving valuable insights from higher levels of abstraction. Operating at elevated levels of abstraction led to limited applicability of formal analysis to the fault injection process. Specifically, our attempts at model learning within the SUMO context resulted in excessively intricate models, often never terminating.

For instance, to learn a faulty model, employing a model learning approach akin to Angluin's $L^*$ algorithm~\cite{Angluin1987} necessitated the establishment of a system \emph{alphabet}. In this context, the alphabet signifies the set of symbols providing context to the system states and transitions. Ensuring that this alphabet adeptly captures both faulty and non-faulty state transitions demanded creative thinking and held a pivotal role in shaping the effectiveness of the acquired model.
 \item{\textbf{The state-space explosion problem:}} 
Both the SAT-based and model learning approaches encountered state space explosion. This challenge emerged due to the logical abstraction and simulation granularity, both significantly influencing the efficiency and effectiveness of the formal analysis. 
A key disparity between fault injection analysis and formal analysis became evident. While fault injection focuses solely on inputs and their corresponding outputs, formal models encompass all inputs and potential outputs, unveiling the system's comprehensive behavior. Consequently, what initially appears as the challenge of examining a finite set of input parameters in fault injection transforms into the state space issue within formal methods. Striking a balance between these two extremes emerges as the sought-after equilibrium. 

\item{\textbf{Non-deterministic behavior:}} SUMO is a deterministic simulator. While constructing the formal model, however, this determinism is lost when augmented with faulty parameters. Additionally, the fault injection community employs randomness within faulty inputs to assess system dependability. This introduces complexity in constructing models that can effectively accommodate such randomness, necessitating techniques like abstraction or alternative formalisms. The non-deterministic aspect might not be immediately evident, potentially yielding unfavorable outcomes if appropriate analysis methods are not selected. 
\end{enumerate}

\section{Conclusions}

In summary, this paper introduces the concept of formal fault injection, an approach that synergizes formal methods and fault injection techniques to enhance the dependability and safety of autonomous systems. By harmonizing development and evaluation phases, this approach facilitates cross-phase knowledge sharing, fostering a unified approach to ensuring dependability attributes. This alignment not only benefits certification bodies and developers but also strengthens the foundation for proving system reliability.

\bibliographystyle{eptcs}
\bibliography{References}
\end{document}